%% file: main-CR.tex
\documentclass[10pt,twocolumn,letterpaper]{article}

\usepackage[pagenumbers]{cvpr} 
\usepackage{times}
\usepackage{epsfig}
\usepackage{graphicx}
\usepackage{amsmath,amsthm}
\usepackage{amssymb}
\usepackage{mathtools}
\usepackage{makecell}
\usepackage{bm,multirow}
\usepackage{multicol}
\usepackage{caption}
\usepackage{subcaption}
\usepackage{comment}

\usepackage{arydshln}

\input{preamble}

\definecolor{cvprblue}{rgb}{0.21,0.49,0.74}
\usepackage[pagebackref,breaklinks,colorlinks,citecolor=cvprblue]{hyperref}

\newcommand{\zb}{{\boldsymbol z}}

\newcommand{\hb}{{\boldsymbol h}}

\newcommand{\epsilonb}{{\boldsymbol \epsilon}}

\usepackage{amsthm}

\def\paperID{14433} 
\def\confName{CVPR}
\def\confYear{2024}

\title{Contrastive Denoising Score for Text-guided Latent Diffusion Image Editing}
\author{Hyelin Nam$^{1}$,\quad Gihyun Kwon$^{2}$,\quad Geon Yeong Park$^{2}$,\quad Jong Chul Ye$^{1}$\\
Kim Jae Chul Graduate School of AI$^1$, Dept. of Bio and Brain Engineering$^2$, KAIST\\
{\tt\small \{hyelin.nam, cyclomon, pky3436, jong.ye\}@kaist.ac.kr}
}

\begin{document}

\twocolumn[{%
\renewcommand\twocolumn[1][]{#1}%
\maketitle
\begin{center}
    \centering
    \includegraphics[width=1.0\linewidth]{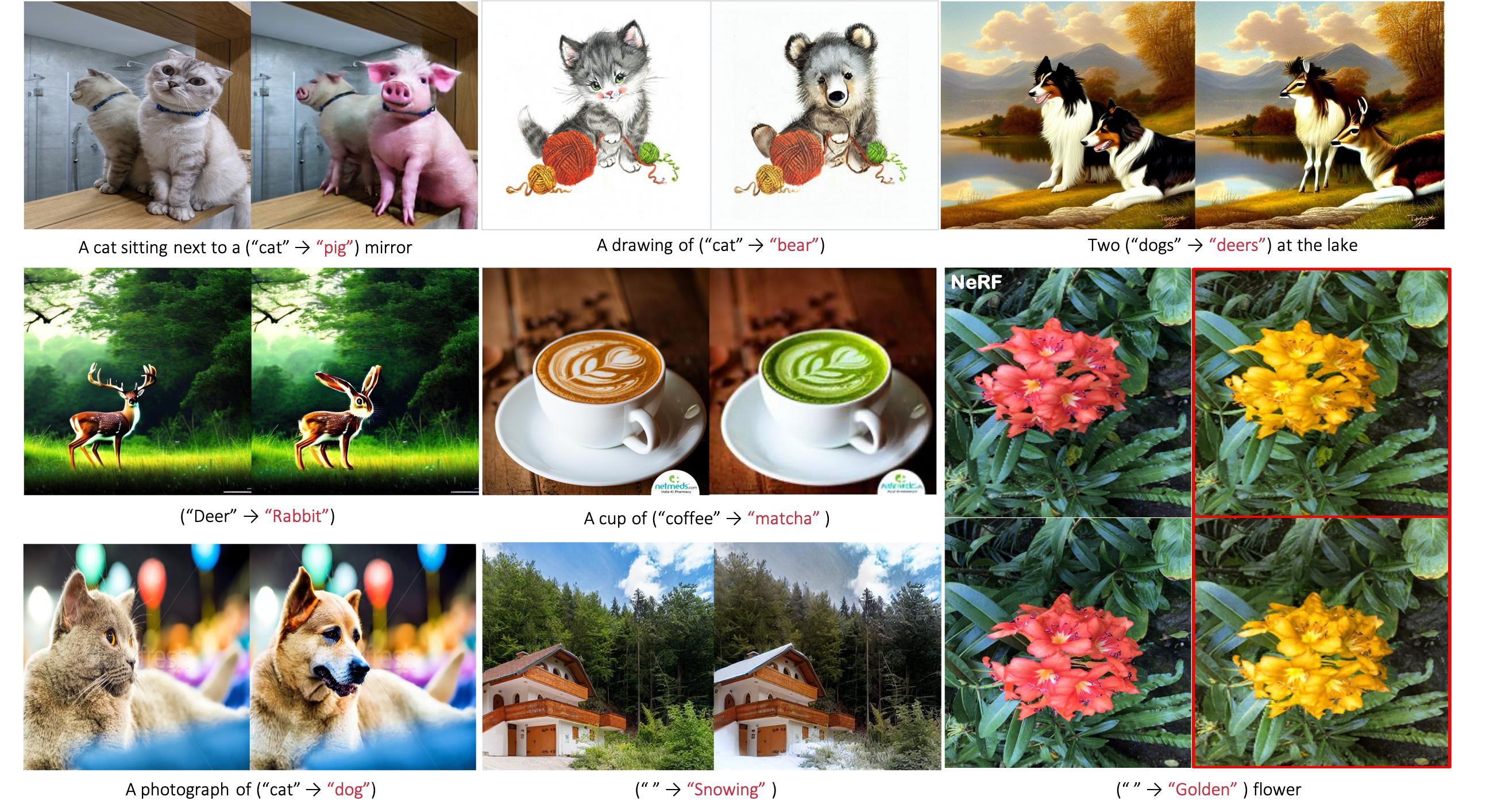}
    \vspace*{-0.6cm}
    \captionof{figure}{Text-guided Image Editing results via \textbf{C}ontrastive \textbf{D}enoising \textbf{S}core. \textbf{CDS} successfully translates source images with a well-balanced interplay between maintaining the structural elements of the source image and transforming the content in alignment with the target text prompt.}
    \label{fig:first}
\end{center}}]

\begin{abstract}
With the remarkable advent of text-to-image diffusion models, image editing methods have become more diverse and continue to evolve. A promising recent approach in this realm is Delta Denoising Score (DDS) - an image editing technique based on Score Distillation Sampling (SDS) framework that leverages the rich generative prior of text-to-image diffusion models.
However, relying solely on the difference between scoring functions is insufficient for preserving specific structural elements from the original image, a crucial aspect of image editing. To address this, here we present an embarrassingly simple yet very powerful modification of
DDS, called Contrastive Denoising Score (CDS), for latent diffusion models (LDM). Inspired by the similarities and differences between DDS and the contrastive learning for unpaired image-to-image translation(CUT), we introduce a straightforward approach using CUT loss within the DDS framework. Rather than employing auxiliary networks as in the original CUT approach, we leverage the intermediate features of LDM, specifically those from the self-attention layers, which possesses rich spatial information. Our approach enables zero-shot image-to-image translation and neural radiance field (NeRF) editing, achieving structural correspondence between the input and output while maintaining content controllability. Qualitative results and comparisons demonstrates the effectiveness of our proposed method. Project page: \url{https://hyelinnam.github.io/CDS/}
\end{abstract}

\section{Introduction}
\label{sec:intro}
\noindent Diffusion models(DMs) have made significant strides in controllable multi-modal generation tasks, particularly in text-to-image(T2I) synthesis. Evolving from basic models, recent Latent Diffusion Model(LDM) showed notable efficacy in T2I task~\cite{rombach2022high,ramesh2022hierarchical,saharia2022photorealistic}. Building on the progress of T2I models, various approaches have been explored to adapt these models for text-conditioned image editing tasks. Initial work primarily focused on incorporating source image conditions into the sampling or reverse process. This involved guiding the generation process through gradient-based sampling~\cite{kwon2023diffusionbased, yang2023zero}, or directly training models that take the source image as conditional input~\cite{brooks2022instructpix2pix}. With the advent of LDM, researchers have begun directly leveraging the properties of T2I models~\cite{tumanyan2023plug,hertz2023prompttoprompt}.

In this context, one promising recent approach is Delta Denoising Score(DDS)~\cite{hertz2023delta} - an image editing technique builds upon the Score Distillation Sampling(SDS)~\cite{poole2023dreamfusion} framework. SDS allows the optimization of a parametric image generator such as 3D NeRF~\cite{mildenhall2021nerf}, capitalizing on the rich generative prior of the diffusion model from which it samples. However, SDS is recognized for prominent concerns such as over-saturation and over-smoothing. To address these issue, DDS introduced an additional reference branch with a matching text prompt to refine the noisy gradient of SDS. Then, DDS queries the generative model with two pairs of images and texts. By utilizing the difference between the outcomes of the two queries, which provides a cleaner gradient direction, the target image is updated incrementally. Unfortunately, in DDS the structural details of the source image are often neglected, as shown in \cref{fig:main}.

As preserving structural consistency is recongized as crucial in image manipulation, mechanisms to enforce structural consistency in the editing process have continued to evolve. In classical CycleGAN~\cite{zhu2017unpaired}, target appearance is enforced using an adversarial loss, while content is preserved using cycle-consistency. However, cycle-consistency assumes the bijection between two domains, necessitating the training of two generator, which can be overly restrictive. Contrastive Unpaired Translation (CUT)~\cite{park2020cut} proposed an alternative approach by maximizing the mutual information between corresponding input and output patches in the latent domains. Building on the success of these approaches, similar ideas have incorporated into diffusion models, utilzing features from ViT~\cite{kwon2023diffusionbased} or the attention layer of the score network~\cite{yang2023zero}. However, these approaches have only been applied to pixel domain diffusion models, not latent diffusion models, and also require additional encoder training, which is inefficient.

To address these challenges, here we propose an embarrassingly simple yet amazingly effective zero-shot training-free method called Contrastive Denoising Score(\textbf{CDS}). This method integrates the application of CUT loss to pretrained latent diffusion models such as Stable Diffusion~\cite{rombach2022high}, within DDS framework. We utilize the rich spatial information in the self-attention features of LDM's to compute the CUT loss. 

Importantly, the application of CUT loss has been limited to pixel-domain diffusion models, and 
we are not aware of any prior work using latent diffusion models. Moreover, while previous image manipulation methods often use attention layers in a supervised manner~\cite{tumanyan2023plug, hertz2023prompttoprompt,parmar2023zero}, our approach utilizes attention in a fully unsupervised manner, randomly selecting sets of patches from the same spatial locations. Therefore, we believe that this is the first work to apply CUT loss to LDM for unsupervised image translation.

We validate the effectiveness of our proposed loss on various text-driven image editing tasks.
Furthermore, as our work is rooted in the score distillation method, it can be applied to various domains, such as Neural Radiance Fields(NeRF)~\cite{mildenhall2021nerf}, as demonstrated in our experiments.

To summarize, we make the following key contributions:
\begin{itemize}
    \item For the purpose of suitable structural consistency, we integrate CUT loss into the DDS framework using latent latent representations. To the best of our knowledge, this is the first work to apply CUT loss LDM for zero-shot image editing.
    \item We demonstrate that utilizing intermediate latent features from self-attention layers enables the application of CUT loss without requiring an additional network. 
    \item We show that our method outperforms existing state-of-the-art baselines, achieving a significantly better balance between preserving the structural details of the original image and transforming the content in alignment with a target text prompt.
    \item This method, being grounded in the score distillation framework, is extendable to multiple domains including NeRF.
\end{itemize}

\begin{figure*}
    \centering
    \includegraphics[width=0.8\linewidth]{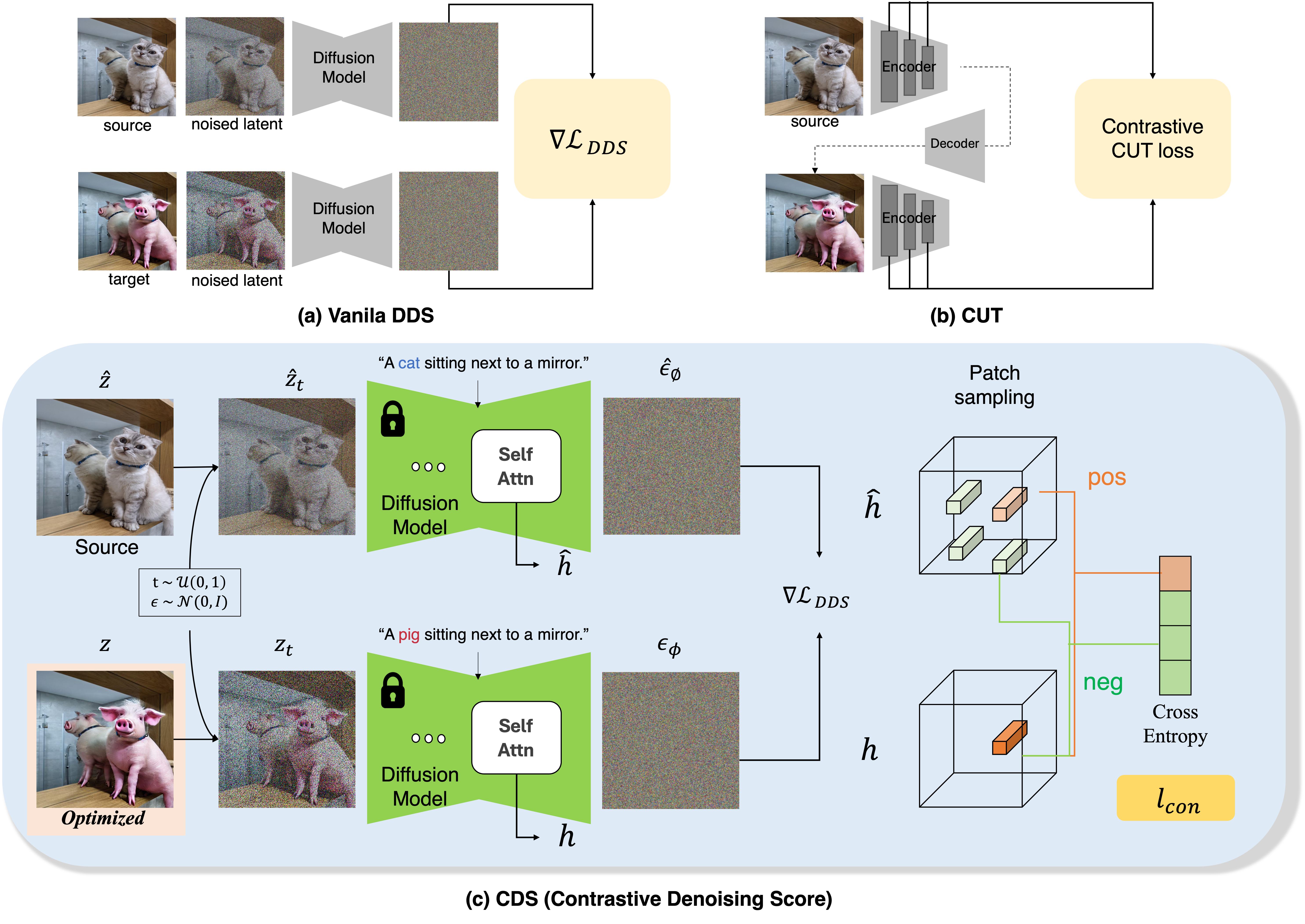}
    \vspace*{-0.3cm}
    \caption{Overall pipeline of \textbf{CDS}. We extract the intermediate features of the self-attention layers and calculate $\ell_{con}$. This loss enables us to regulate structural consistency and generate reliable images.}
    \label{fig:method}
\end{figure*}

\section{Related works}
\noindent\textbf{Image editing with Text-to-image Diffusion Models} Recently, the advancements in Text-to-image Diffusion models have led to generated images that closely resemble real-world visuals, often indistinguishable to the human eye~\cite{rombach2022high,saharia2022photorealistic,ramesh2022hierarchical}. The popularity of open-source generative models, particularly models like LDM~\cite{rombach2022high}, has led to extensive exploration in image editing. The incorporation of a text-conditioned injection framework through the cross-attention layer of the model has enabled a diverse image editing~\cite{hertz2023prompttoprompt} 
and translation~\cite{tumanyan2023plug,meng2022sdedit} tasks. 
Recent methods also have introduced innovative approaches such as re-weighting for editing specific components ~\cite{hertz2023prompttoprompt}, the injection of self-attention features for image translation ~\cite{tumanyan2023plug}, and combinations with methods for inverting real images~\cite{mokady2022nulltext,han2024proxedit}. These approaches demonstrate improved editing performance. 

While most of these methods are typically applied during the reverse process in pre-trained models, an extended method known as Score Distillation Sampling (SDS)~\cite{poole2023dreamfusion} method has shown promising performance, both in 3D object generation and 2D image generation tasks. However, SDS often leads to blurry outputs due to its reliance on the gradient of the difference between pure noise and the target text score prediction. To address this, Delta Denoising Score (DDS)~\cite{hertz2023delta} introduces an alternative editing approach using the gradient between the source text score and the direction of the target text score. Despite this improvement, DDS overlooks the critical aspect of editing: maintaining structural consistency between the source and output images, thus limiting its overall editing performance. To address this, we propose a new framework that enhances the performance of DDS by introducing an appropriate contrastive loss to maintain structural
similarity.

\vspace{-0.3cm}
\paragraph{Consistency Regularization for Image Manipulation}
In image editing and translation, preserving the structural components of the source image while transforming its semantics is crucial. The initial work of 
CycleGAN~\cite{zhu2017unpaired} introduced cycle consistency, translating output images back to the source domain. Building on this, subsequent studies~\cite{fu2019geometry,huang2018multimodal,benaim2017one} proposed various consistency regularization techniques to enhance Image-to-Image (I2I) performance. On the other hand, Contrastive Unpaired Translation (CUT)~\cite{park2020cut} introduced a method of applying contrastive learning to patch-wise representations, effectively preserving structural information between source and output. Inspired by this work,  subsequent studies have explored the application of the CUT loss to StyleGAN~\cite{kwon2023one} and Diffusion models~\cite{kwon2023diffusionbased, yang2023zero}, yielding promising results. While various techniques have been proposed, applying these methods to pretrained text-to-image latent diffusion models like StableDiffusion still remains an open problem, as fine-tuning the off-the-shelf LDM is computationally intensive. We discovered a natural integration of CUT loss into the off-the-shelf LDM within the DDS framework without fine-tuning. This zero-shot integration significantly improves editing output compared to traditional DDS approaches.

\section{Main Contribution: The CDS}
\label{sec:formatting1}

\subsection{Key Observation}

\noindent\textbf{DDS.} We begin with a concise overview of DDS and then compare its similarities and differences with CUT. This comparative explanation clearly illustrates the missing component in DDS and inspires us on how to improve DDS.

The noise that text conditioned diffusion models using classifier-free guidance (CFG)~\cite{ho2021classifierfree} predicted can be expressed as:
\begin{equation}
\epsilonb_{\phi}^{\omega}(\zb_{t}, y, t) = (1+{\omega}) \epsilonb_{\phi}(\zb_{t}, y, t) - \omega \epsilonb_{\phi}(\zb_{t}, \emptyset, t)
  \label{eq:CFG}
\end{equation}
where $\omega$ denotes the guidance parameter, and
${\epsilonb}_{\phi}$ denote a noise prediction network with parameters set ${\phi}$.
Additionally,
$y$ and $\emptyset$ represent the text and null-text prompt, respectively,
and while $\zb_t$ represent a noisy latent from the clean latent $\zb_0$ at the noise timestep ${t}{\sim}{\mathcal{U}}({0,1})$.

SDS leverages the gradient of the diffusion loss function:
\begin{align}
\mathcal{L}_{\mathrm{SDS}}(\theta;y_{tgt})=\|\epsilonb_{\phi}^{\omega}(\zb_t(\theta), y, t) - \epsilonb\|^2
\end{align}
where $y$ refers to the target prompt,
 ${\epsilonb}{\sim}{\mathcal{N}}({0,\mathbf{I}})$ and
\begin{align}
\zb_t(\theta)=a_t \zb_0(\theta)+ b_t
\end{align}
for some DDPM~\cite{ho2020denoising} noise schedule $(a_t,b_t)$ with $a_t^2+b_t^2=1$ and
with $\zb_0(\theta)$ denotes the target latent parameterized by $\theta$ that should be optimized to follow the target prompt. It has been demonstrated that $\nabla_{\theta}\mathcal{L}_{\mathrm{SDS}}$ is an efficient gradient term for generating images that exhibit a heightened level of fidelity to the given prompt. 
However, SDS produces blurry outputs that primarily emphasize objects mentioned in the prompt, rendering it insufficient for practical image editing purposes.

DDS expands the SDS framework for image editing, utilizing not only the target text prompt $y$ but also a reference pair of image $\hat{\zb}_0$ and text $\hat{y}$. Specifically, the DDS loss is given by
\begin{align}\label{eq:lossDDS}
\mathcal{L}_{\mathrm{DDS}}(\theta;y_{tgt})=\|\epsilonb_{\phi}^{\omega}(\zb_t(\theta), y, t) - \epsilonb_{\phi}^{\omega}(\hat\zb_t, \hat y, t)\|^2
\end{align}
where $\hat \zb_t = a_t \hat \zb_0+b_t$. In the same manner as SDS, $\zb_0(\theta)$ is updated incrementally in the direction of $\nabla_{\theta}\mathcal{L}_{\mathrm{DDS}}$. 
As shown in ~\cref{fig:method}a, 
the score supplied by the reference branch aligns with the score
from the output branch by minimizing \eqref{eq:lossDDS}. 
Given that
the score $\epsilonb_{\phi}^{\omega}$ can be understood as a feature vector, DDS aims to minimize the differences in the feature domain.

\begin{figure*}
    \centering
    \includegraphics[width=0.8\linewidth]{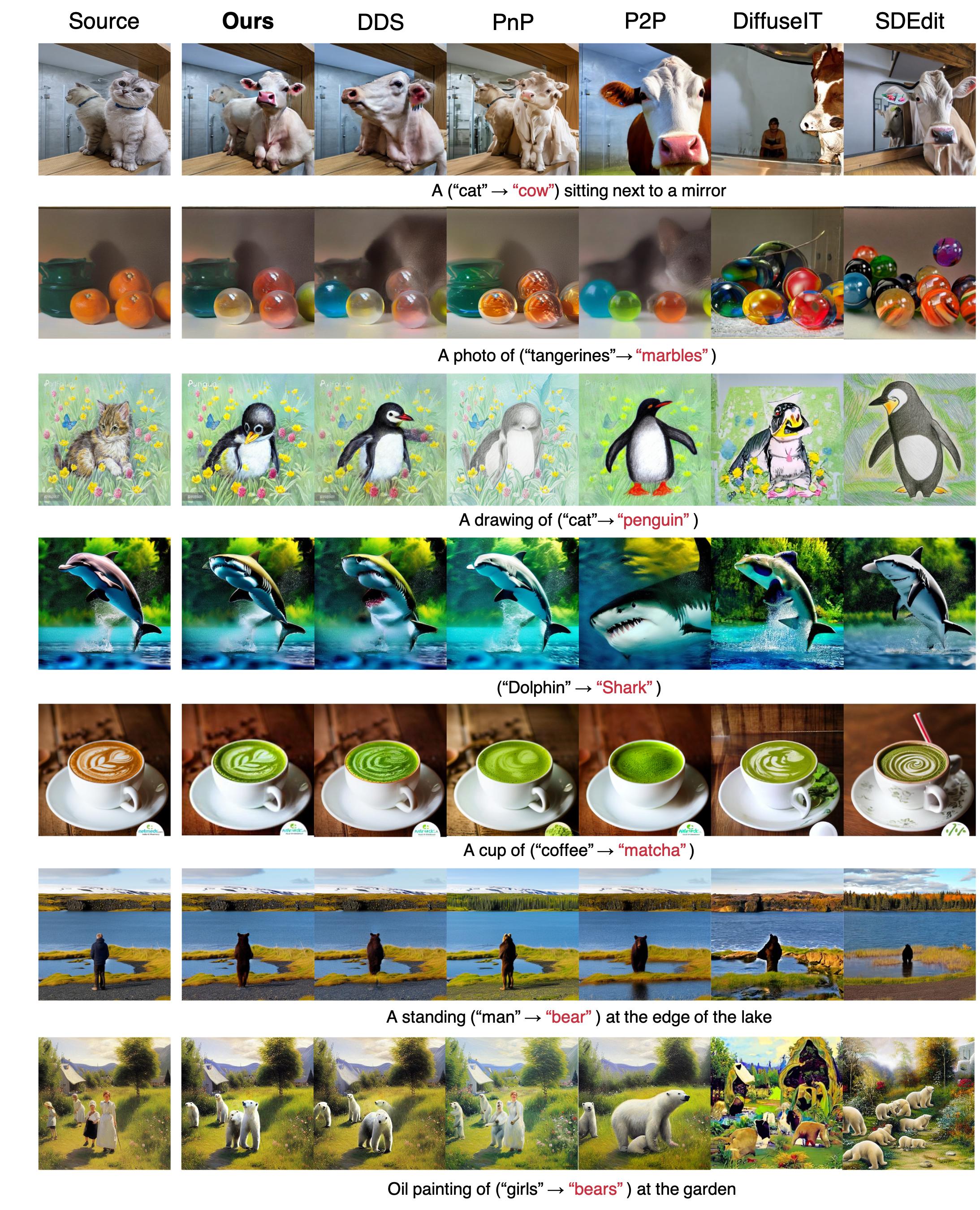}
    \vspace{-0.1cm}
    \caption{Comparison with baseline models. CDS demonstrates outstanding performance in effectively regulating structural consistency.}
    \label{fig:main}
\end{figure*}

\noindent\textbf{CUT.}
The fundamental idea of CUT is to exploit
patch-wise contrastive learning in the feature domain for one-sided translation. Specifically, the generated 
output should produce features whose patches resemble corresponding patches from the input image, in contrast to other random patches. As shown in ~\cref{fig:method}b, CUT utilizes a multilayer, patch-wise contrastive loss, which
maximizes mutual information between corresponding input and output patches. This facilitates one-sided translation in the unpaired setting without imposing the need for enforcing the cycle consistency. CUT has been demonstrated that the CUT loss is effective in maintaining correspondence in content and structure, making it widely utilized technique in image editing.

\subsection{Contrastive Denoising Score}
By inspection of~\cref{fig:method}a and~\cref{fig:method}b,
we observe a striking similarity between the two approaches. Aside from the image generation processes (i.e., via a trained decoder network in CUT and image optimization in DDS),
both approaches aim to align the features from the reference
and reconstruction branches. 

While DDS offers a denoised editing direction that focuses on editing the pertinent part of the image to match the target text, empirical observations reveal instances of failure cases. For example, as shown in~\cref{fig:main}, the pose or structural details of the content in the source image are not preserved. Recall that the main objective of text-driven image editing is not only aligning to the content specified in a target prompt, but also incorporating the structure and details of an input source image. 

With the aim of regulating excessive structural changes, we are interested in the key idea from CUT, which has been shown to be effective in maintaining the input structure. However, the original CUT algorithm requires training an encoder to extract spatial information from the input image, which is inefficient. Therefore, we aim to calculate CUT loss without introducing an auxiliary encoder, fully leveraging the information of the latent representation of LDM. 

One potential approach is to compute CUT loss by directly leveraging the scores 
$\epsilonb_{\phi}^{\omega}(\zb_t(\theta), y, t)$ and $\epsilonb_{\phi}^{\omega}(\hat\zb_t, \hat y, t)$. While this approach demonstrates effectiveness, in certain cases, we have 
observed that  semantic changes to align with the content specified in a target text were also suppressed due to information entanglement. On the other hand, intermediate representations from self-attention layers have been shown to contain rich spatial information, disentangled from semantic information~\cite{tumanyan2023plug}. Even considering how it operates, self-attention layers contain similarity information between spatial patches in the given representations, which is exactly CUT loss requires. Therefore, we calculate CUT loss utilizing the latent representation of self-attention layers. 
For visualization results of CUT loss in other feature extraction spaces, please refer to ~\cref{fig:ablation_cut}.

Specifically, we begin by briefly describing the self-attention layers that compose the denoising U-Net $\epsilonb_{\theta}$ in the Stable Diffusion (SD) model. During each timestep $t$ of the denoising process, the noisy latent representation $\zb_t$ is fed as input to the denoising network. For self-attention layer $l$, $\hat{\hb_l}$ and $\hb_l$ represent the intermediate features passed through the residual block and self-attention block conditioned on $\hat{y}$ and $y$, respectively.

During the denoising process of each branch, which is the part of the DDS gradient computation, we obtain $\hat{\hb}_l$ and $\hb_l$. Then, we randomly select patches from the feature map ${h_l}$. Initially, a ``query" patch is sampled from the feature map ${\hb_l}$. We denote $s \in \left\{1, ..., S_l\right\}$, where $S_l$ is the number of query patches. Then, for each query, the patch at the corresponding location on the feature map $\hat{\hb_l}$ is designated as the ``positive", while the non-corresponding patches within the feature map serve as ``negatives". We refer to the positive patch as $\hat{\hb^s_l}$ and the other patches as $\hat{\hb_l}^{S\setminus s}$. The objective of the CUT loss is to maximize the mutual information between ``positive" patches while simultaneously minimizing the mutual information between ``negatives."  This process can be formulated as the patchNCE loss:
\begin{equation}
\ell_{con}({\zb}, \hat{\zb}) = \mathbb{E}_{\hb}\left [ \sum_{l} \sum_{s} \ell({\hb^s_l},\hat{\hb^s_l}, \hat{\hb}_l^{S\setminus s})  \right ]
  \label{eq:L_con} 
\end{equation}
where, $\ell\left ( \cdot  \right )$  denotes cross-entropy loss:
\begin{align}
\ell(\hb,\hb^+,\hb^-)= -\log\left(\frac{\exp(\hb\cdot\hb^+/\tau)}{\exp(\hb\cdot\hb^+/\tau)+\sum \exp(\hb\cdot\hb^-/\tau)}
\right)
\end{align}
for some parameter $\tau>0$.
By additionally incorporating this simple $\ell_{con}$, we can regularize DDS to maintain structural consistency between $\zb$ and $\hat{\zb}$.

\begin{table}[t]
    \centering
    \resizebox{1.0\columnwidth}{!}{
    \begin{tabular}{c|c c c |c c c}
        \toprule
        \multirow{2}{*}{Method} & \multicolumn{3}{c|}{\textbf{cat2dog}} & \multicolumn{3}{c}{\textbf{cat2cow}}
        \\
         & CLIP Acc ($\uparrow$) & Dist ($\downarrow$) & LPIPS ($\downarrow$) & CLIP Acc ($\uparrow$) & Dist ($\downarrow$) & LPIPS ($\downarrow$)
         \\
        \midrule
        {SDEdit + word swap} & \textbf{{97.9$\%$}} & {0.052} & {0.138} & {97.9$\%$} & {0.052} & {0.138}\\
        {DiffuseIT} & \textbf{{97.9$\%$}} & {0.174} & {0.277} & 
        {93.3$\%$} & {0.175} & {0.302} \\
        {DDPM inv + P2P} & {86.1$\%$} & {0.072} & {0.150} &
        {86.1$\%$} & {0.078} & {0.158} \\
        {DDPM inv + PnP} & {91.6$\%$} & {0.073} & {0.156} &
        {95.0$\%$} & {0.079} & {0.179} \\
        {DDS} & \textbf{{97.9$\%$}}  & {0.023} & {0.080} &
        \textbf{{99.6$\%$}} & {0.040} & {0.116} \\
        {\textbf{Ours}} & {97.5}$\%$ & {\textbf{0.020}}& {\textbf{0.079}} &
        {97.9$\%$} & \textbf{0.033} & \textbf{0.112} \\
        \bottomrule
    \end{tabular}
    } 
    \caption{Quantitative evaluation for the cat $\rightarrow$ dog and cat $\rightarrow$ cow tasks. 'Dist' denotes DINO-ViT structure distance. Results for the cat $\rightarrow$ pig are provided in our Supplementary Materials.}
    \vspace{-0.1cm}
    \label{table: quantitative evaluation}
\end{table}

\begin{table}[h]
\centering
\resizebox{0.8\columnwidth}{!}{
\begin{tabular}{c c c c}
\toprule
\multicolumn{1}{c}{\multirow{2}{*}{\textbf{Method}}} & \multicolumn{3}{c}{\textbf{Metric}} \\
\cmidrule{2-4} 
\multicolumn{1}{c}{} & Text ($\uparrow$) & Structure ($\uparrow$) & Quality ($\uparrow$)\\
\midrule
{SDEdit + word swap} & {3.77} & {2.90} & {3.43} \\
{DiffuseIT} & {3.17} & {2.94} & {2.83}\\
{DDPM inv + P2P} & {3.89} & {2.69} & {3.61}  \\
{DDPM inv + PnP} & {3.36} & {3.70} & {3.22}\\
{DDS} & {4.06} & {4.05} & {3.64} \\
{\textbf{Ours}} & {\textbf{4.43}} & {\textbf{4.65}} & {\textbf{4.20}} \\
\bottomrule
\end{tabular}
}
\caption{User study results.}
\label{table: user study}
\end{table}

\section{Experiments}
\subsection{Experimental setting}
\noindent\textbf{Implementation.} For implementation, we referenced the official source code of DDS\footnote{\url{https://github.com/google/prompt-to-prompt/blob/main/DDS_zeroshot.ipynb}} by using Stable Diffusion v1.4. More details are provided in our Supplementary Materials. 

\noindent\textbf{Baseline methods.}
To comprehensively evaluate the performance of our method, we conduct comparative experiments, comparing it to several state-of-the-art methods.
Our method is compared against five baselines including vanilla DDS. For implementation, we referred the official source code of each methods, except for SDEdit~\cite{meng2022sdedit} which we used the implementation of Stable Diffusion. For methods requiring additional inversion process, we employed DDPM inversion ~\cite{huberman2023edit}.

\subsection{Experimental Results}
\paragraph{Qualitative Results.} To compare the qualitative results, we show the edited outputs in~\cref{fig:main}.
For the sampling-based methods of DiffuseIT and SDEdit, the results change the source image attribute to target text conditions. However, the output structures are severely deformed compared to the original source image structures, and in some cases, unwanted artifacts are present. For the attention-based method of Prompt-to-Prompt(P2P)~\cite{hertz2023prompttoprompt}, the results also follow the target text conditions. However, the method suffers from severe structure inconsistencies because the quality is largely affected by the performance of the inversion method and the timesteps for attention map modulation. For the Plug-and-Play(PnP)~\cite{tumanyan2023plug} diffusion baseline, and the overall structure is well maintained from source images, and the output semantics reflect the text conditions. However, the method still does not fully maintain consistency between source and output, and for difficult cases such as cat$\rightarrow$cow, the output shows unrealistic result with artifacts. DDS shows decent performance in text-guided editing, but still it fails to maintain structural consistency between the source images. On the other hand, our method, CDS successfully edits the source images while preserving their original structural information and does not modify the unrelated regions (e.g. background).

\begin{figure}[t]
    \centering
    \includegraphics[width=0.95\linewidth]{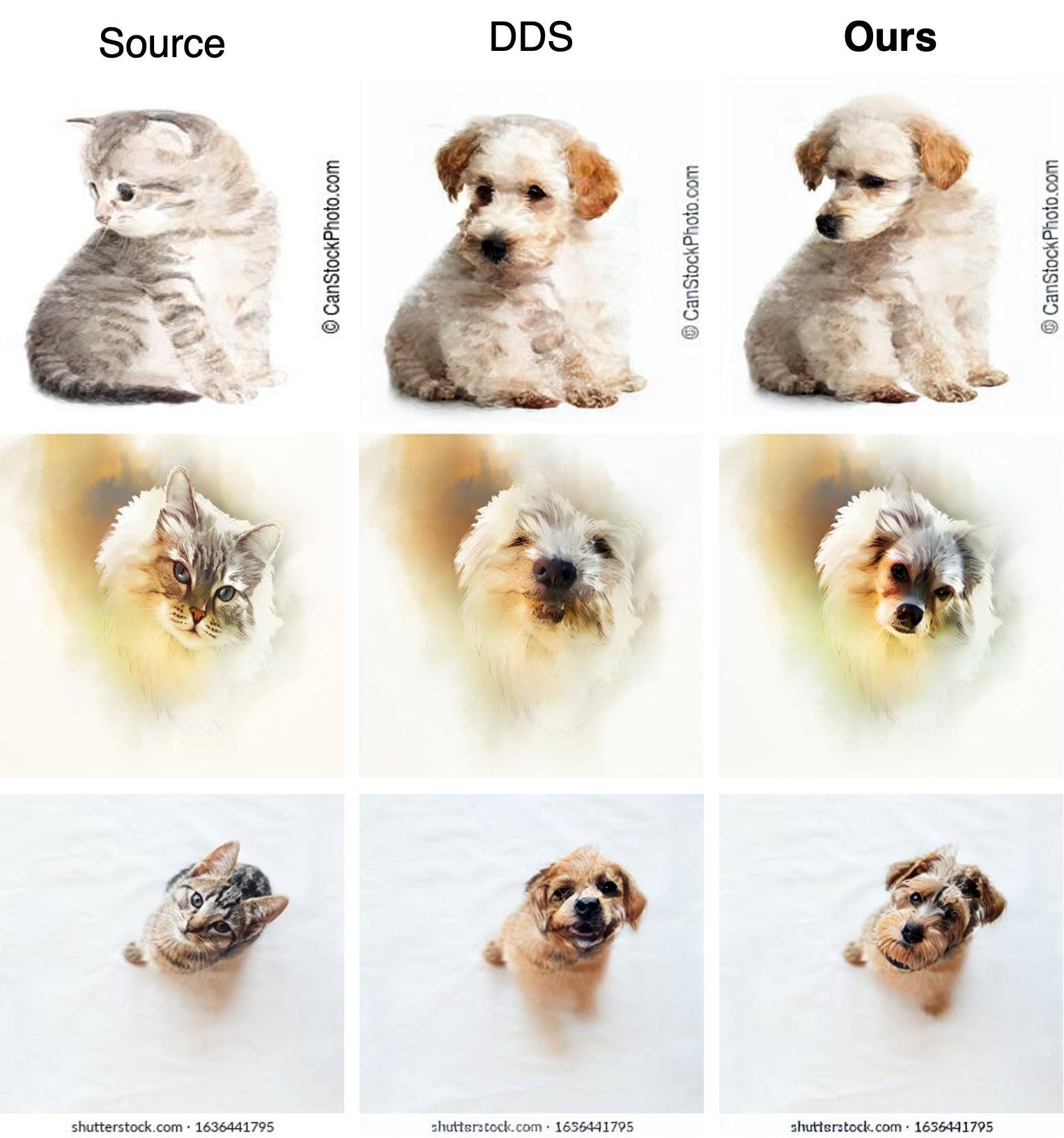}
    \caption{Sample results of the cat2dog task from DDS and CDS.}
    \label{fig:quantitative_comparison}
    \vspace{-0.3cm}
\end{figure}

\paragraph{Quantitative results} In order to further measure the generation quality of our proposed method, we conduct three tasks: (1) cat $\rightarrow$ dog, (2) cat $\rightarrow$ cow and (3) cat $\rightarrow$ pig. Adhering to the data collection protocol outlined in ~\cite{parmar2023zero}, we gather 250 images relevant to cats from the LAION 5B dataset~\cite{schuhmann2022laion5b}. In order to assess whether targeted semantic contents are accurately reflected in the generated images, we measure CLIP~\cite{radford2021learning} accuracy(CLIP Acc). Additionally, to measure the structural consistency, we calculated DINO-ViT~\cite{caron2021emerging} structure distance and  LPIPS distance between the given source image and the outputs. The DINO-ViT structure distance is defined as the difference in self-similarity among the keys obtained from the attention module at the deepest layer of DINO-ViT. ~\cref{table: quantitative evaluation} shows that our proposed method achieves high CLIP-Acc while maintaining low structure distance, indicating optimal editing results while preserving the structural elements of the original input image. Visualizations are provided in the Figure \ref{fig:quantitative_comparison}.

For human evaluation, we presented six comparison results and collected feedback from 20 participants. After each participant viewed images generated by our model and the baselines, they provided feedback through a scoring survey. We set the minimum score as 1 and the maximum score as 5, and user choose the score among 5 options. The following are the provided questions for evaluation: (1) (Text-match) \emph{Does the image reflect the target
text condition?} (2) (Structural consistency) \emph{Does the image contain the content and structure information of source images?} (3) (Overall quality) \emph{Are the generate images realistic?} In ~\cref{table: user study}, our model showed the best performance. 

\begin{figure}[t]
    \centering
    \includegraphics[width=0.9\linewidth]
    {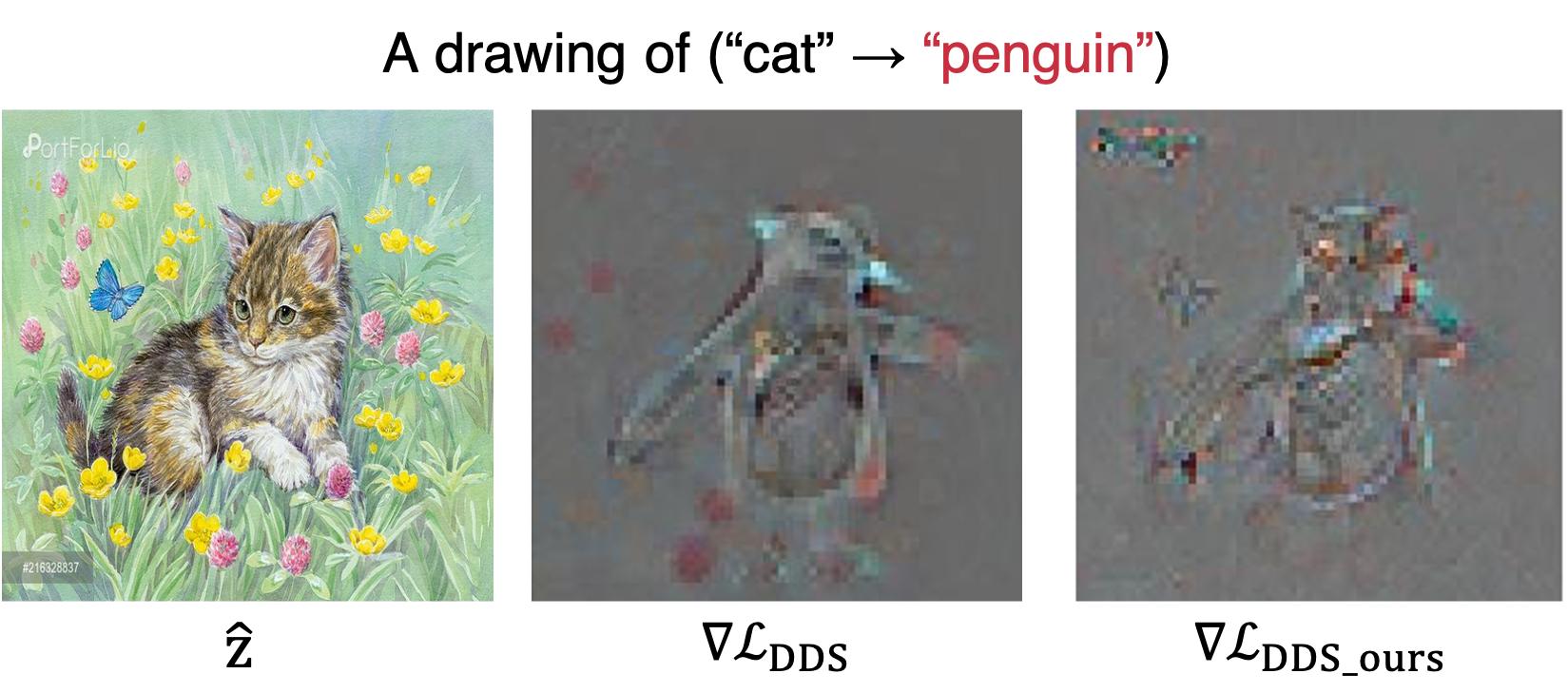}
    \vspace{-0.3cm}
    \caption{Gradient visualization on DDS and CDS.}
    \label{fig:ablation_gradient}
    \vspace{-0.3cm}
\end{figure}

\section{Analysis}
\subsection{Ablation Study}
\paragraph{CUT loss.} First, we ablate our loss $\ell_{con}$ and demonstrate its effectiveness in \cref{fig:quantitative_comparison}. We observed that excluding $\ell_{con}$, which is vanilla DDS, resulted in a loss of structural details even though the overall contents are changed. When we apply $\ell_{con}$, structural attributes such as leg and pose are preserved. This indicates that $\ell_{con}$ is beneficial for preserving the overall structure of source image. Overall, when we apply $\ell_{con}$ using features from the self-attention layers, we can reliably edit images while reflecting both the source image structure and the target text semantics.

Furthermore, we also conducted a study on the visualization of gradients for both vanilla DDS and CDS with the proposed loss. In~\cref{fig:ablation_gradient}, we present the visualization results of gradients. For the gradients of vanilla DDS, the spatial information does not accurately follow the original structure of the source images. However, in CDS, the gradient contains much more detailed structural information, such as cat's ears and pose. This shows that our proposed CUT loss framework enhances the spatial details of the DDS gradient, leading to further improvement in the final edited output.

\begin{figure}[t]
    \centering
    \includegraphics[width=1.0\linewidth]
    {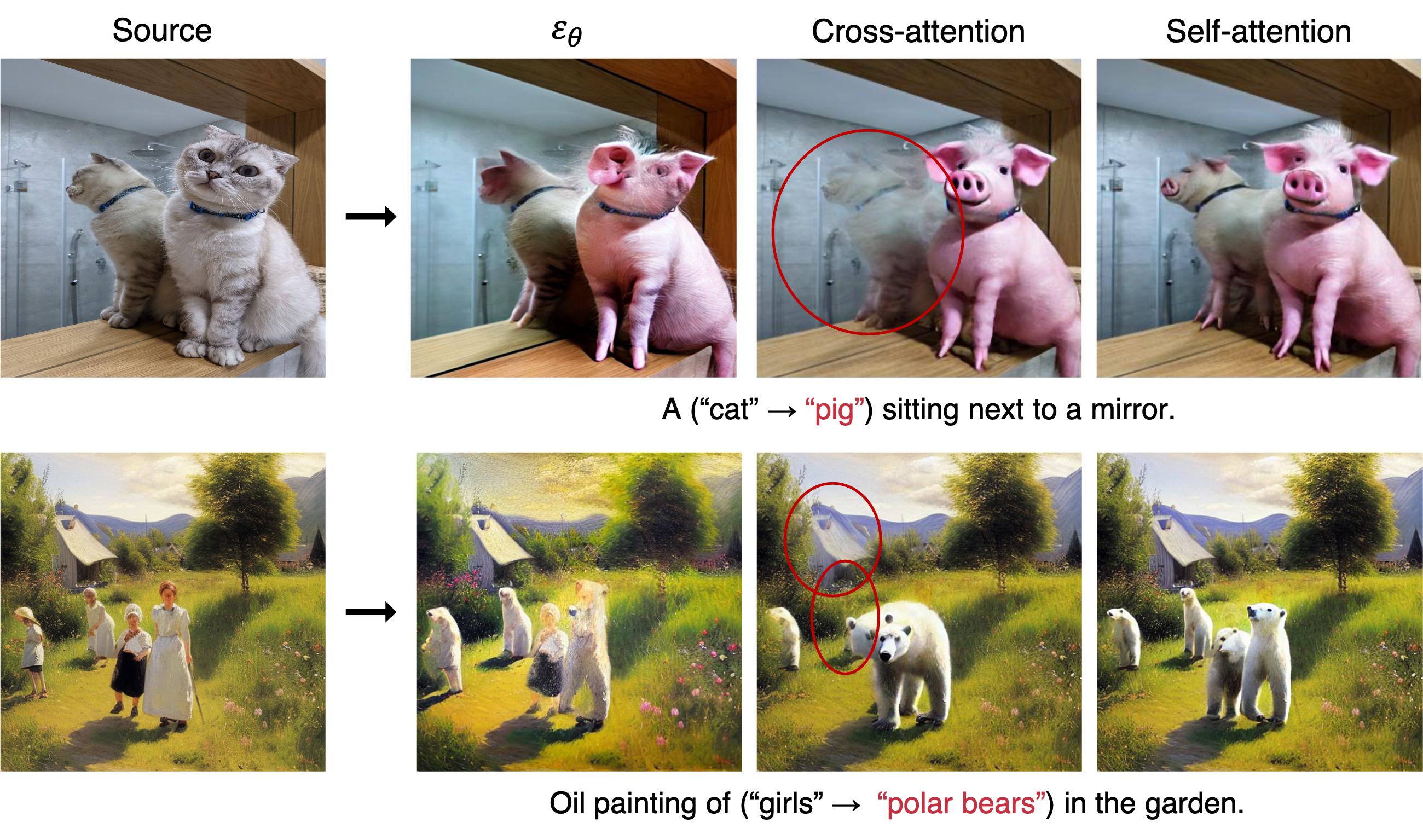}
    \vspace{-0.5cm}
    \caption{Qualitative results for ablation study on feature extraction location for  contrastive loss.}
    \label{fig:ablation_cut}
    \vspace{-0.2cm}
\end{figure}

\paragraph{CUT loss location.}
First, we evaluated the effectiveness of feature extraction layer for CUT loss calculation, as shown in ~\cref{fig:ablation_cut}. We show the generated outputs with the additional CUT loss applied on the direct score network output, the hidden state of cross attention layer, and the hidden state of self-attention layers, respectively.
Additionally, the results in~\cref{fig:ablation_cut} illustrate that directly applying the losses to the score output excessively constrains information. When applying the loss on cross attention features, we can see that the outputs do not correctly preserve the original structures. Our results show that applying proposed method has the best performance. This demonstrates that the features extracted from self-attention layers possess disentangled spatial information, which aligns well with the requirements of CUT loss.

\begin{figure}[t]
    \centering
    \includegraphics[width=1.0\linewidth]
    {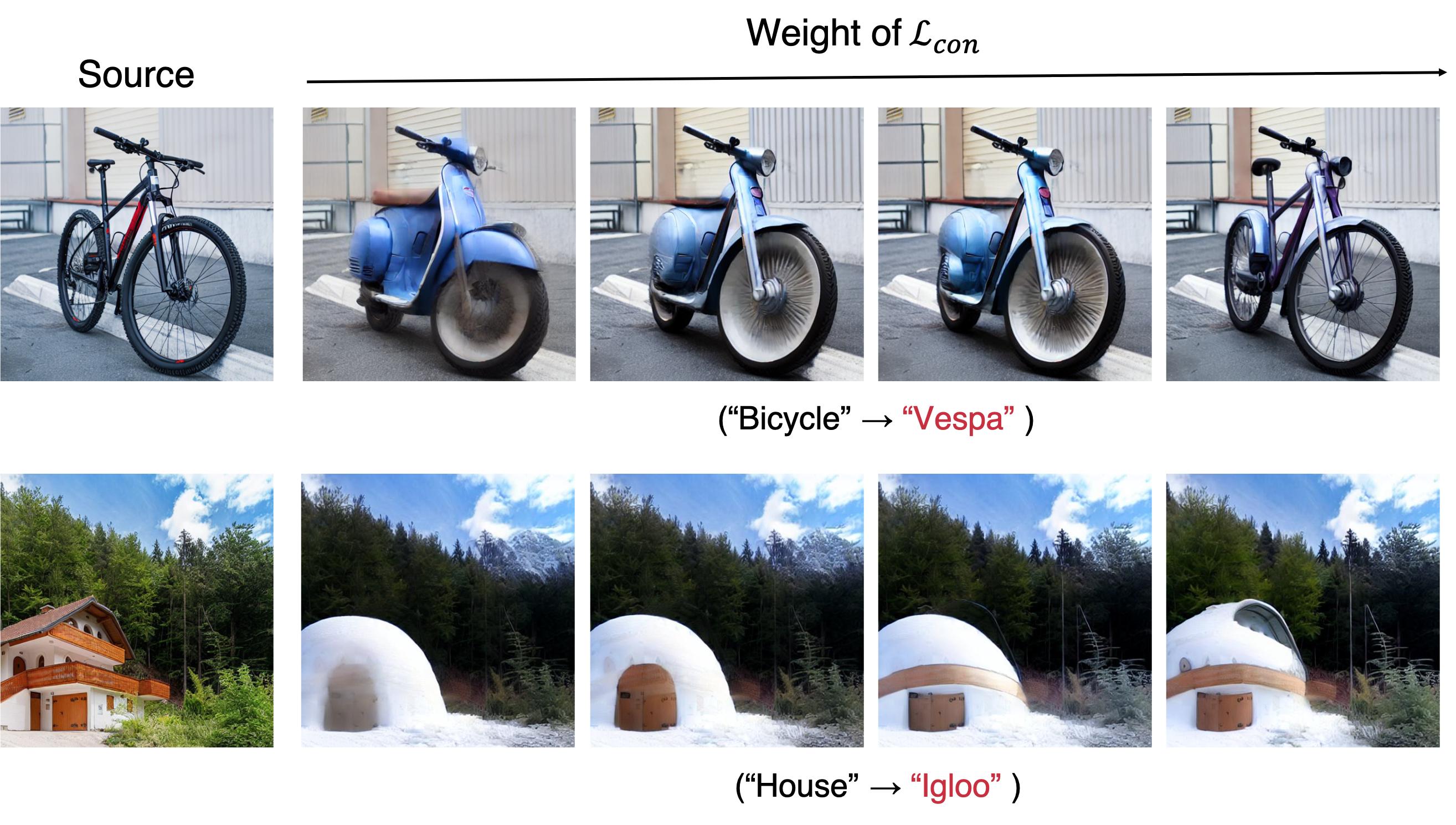}
    \vspace{-0.5cm}
    \caption{Ablation study on weights of contrastive loss.}
    \label{fig:ablation_w}
    \vspace{-0.2cm}
\end{figure}

\begin{figure}[t]
    \centering
    \includegraphics[width=1.0\linewidth]{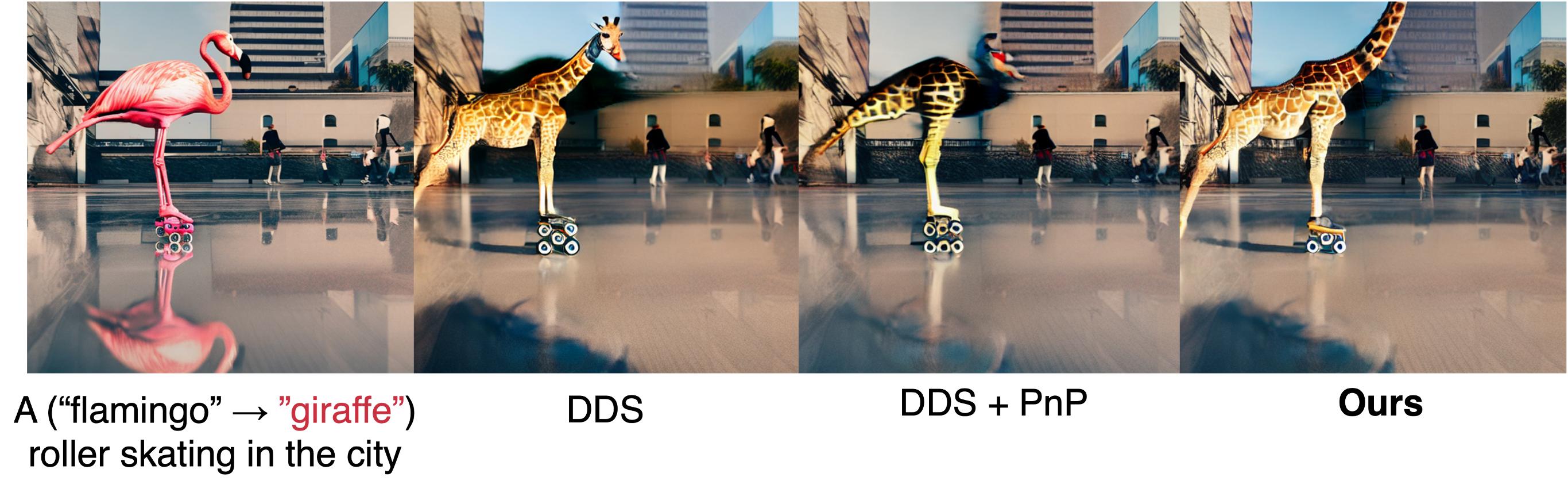}
    \vspace{-0.5cm}
    \caption{Comparison with feature injection method.}
    \label{fig:pnp_comparison}
\end{figure}

\paragraph{CUT loss weight.}
For further analysis of our proposed loss, we conducted additional experiments by varying the loss weights. Intuitively, we expect better structural consistency with higher contrastive loss weights and better semantic changes with lower weights. In~\cref{fig:ablation_w}, we observe that we can control the editability and consistency by varying the weight of contrastive loss. We can see that using a stronger loss not only affects the preservation of the target object structure but also impacts the preservation of the background area.

\subsection{Comparison with feature injection method}
We noted that recent works, including PnP, leverage feature injection for the purpose of structure preservation. Therefore, we compared the outputs of DDS, DDS with feature injection following the PnP approach, and CDS. In~\cref{fig:pnp_comparison}, we can see that while feature injection aids in structural preservation, its forced application without considering semantic information can result in over-constraint during extensive edits. In contrast, CDS employs regularization based on semantic similarity between patches, determined through the inner product in a well-pretrained feature space. This enables CDS to adeptly balance text-aligned editing with the preservation of structural elements.

\begin{figure}[t]
    \centering
    \includegraphics[width=1.0\linewidth]{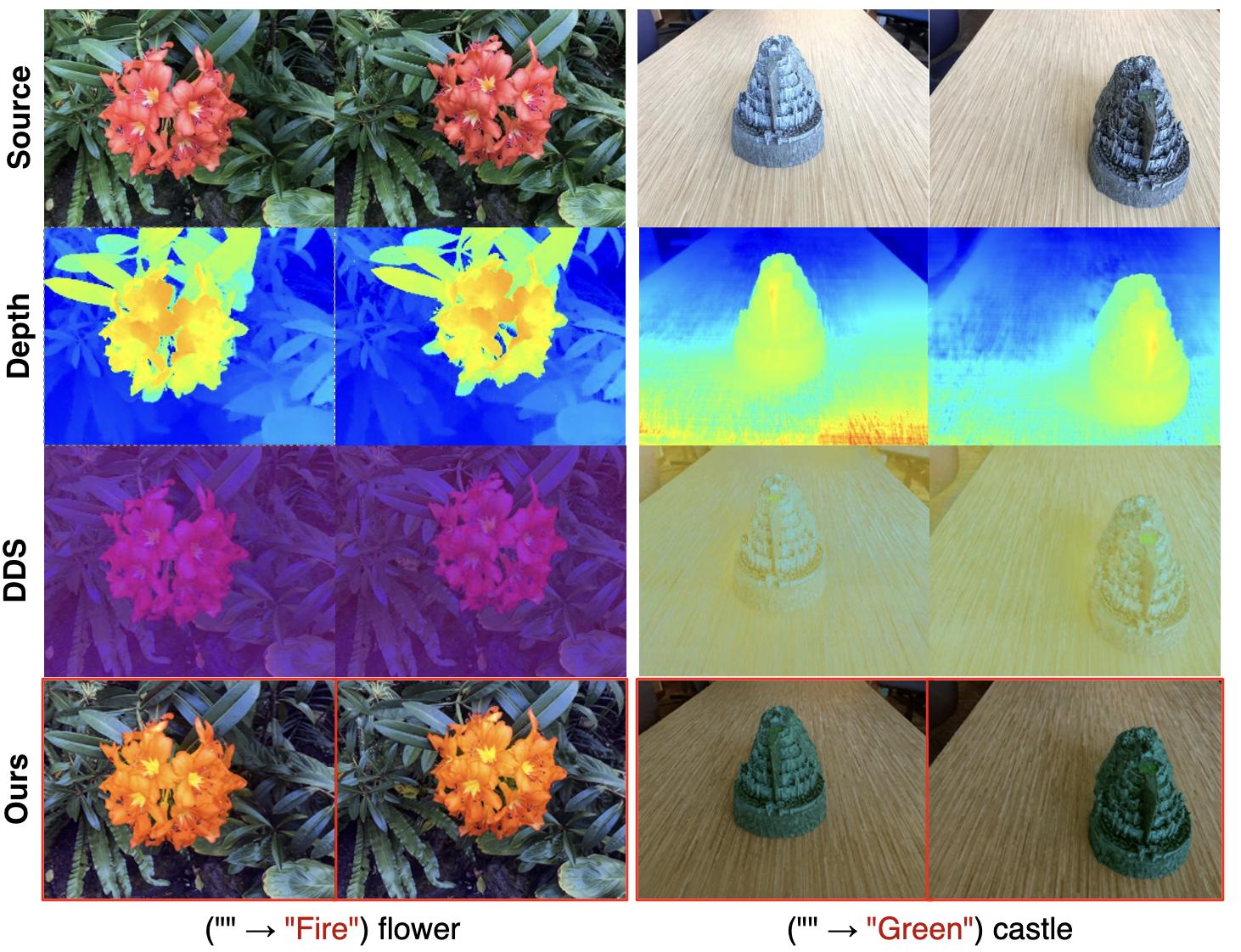}
    \vspace{-0.3cm}
    \caption{Results on NeRF editing. As an extension of our proposed framework, we applied our method to the NeRF 3D object editing task.}
    \label{fig:nerf} 
    \vspace{-0.1cm}
\end{figure}

\subsection{Extension to NeRF}
Since our propose method is an improved version of the score distillation framework, we can transfer the distilled score gradient to other generator networks, such as NeRF. Inspired by the recent work ED-NeRF~\cite{park2024ednerf}, which aimed to leverage DDS for text-guided 3D object editing, we applied our proposed method to a pre-trained NeRF. Beginning with original NeRF model, we rendered the 2D images for reference and target NeRF models with the same view directions. Subsequently, we applied our proposed framework to the NeRF fine-tuning task. For comparison, we also conducted experiments with our baseline DDS. Additional experimental details are provided in our Supplementary Materials.

In \cref{fig:nerf}, we compare the results of applying vanilla DDS and our proposed framework to pre-trained NeRF model. With vaniila DDS, the outputs fail to capture the structural information of source images, resulting in changes to entire pixel colors. However, our proposed method accurately edits the target object(e.g.a flower) to match the desired text conditions. These results demonstrate the effectiveness of our proposed scheme in the 3D NeRF space, highlighting that our method is extendable to multiple domains based on the score distillation framework.

\section{Conclusion}
\quad In this paper, we proposed CDS, which incorporates the CUT loss within DDS framework to preserve structural consistency. Unlike the original CUT algorithm, which required additional network inference, we leverage the rich spatial information inherent in latent representations extracted from LDM, particularly the self-attention layer. This loss allows us to successfully generate images with a better balance between preserving the structural details of the original image and transforming the content in alignment with a target text prompt. Qualitative and quantitative experiments demonstrate that the effectiveness of our proposed method and its scalability to multiple domains. Regarding limitations, failure cases can arise from unfavorable random patch selections or when the source object has unconventional poses, such as looking back.

\section*{Acknowledgements}
This work was supported by Institute of Information \& communications Technology Planning \& Evaluation (IITP) grant funded by the Korea government(MSIT)  
(No.2019-0-00075, IITP grant funded by the Korea government(MSIT, Ministry of Science and ICT) (No.2022-0-00984, Development of Artificial Intelligence Technology for Personalized Plug-and-Play Explanation and Verification of Explanation), Culture, Sports and Tourism R\&D Program through the Korea Creative Content Agency grant funded by the Ministry of Culture, Sports and Tourism in 2023, and Field-oriented Technology Development Project for Customs Administration funded by the Korea government (the Ministry of Science \& ICT and the Korea Customs Service) through the National Research Foundation (NRF) of Korea under Grant NRF2021M3I1A1097910 \& NRF2021M3I1A1097938.

{
    \small
    \bibliographystyle{ieeenat_fullname}
    \bibliography{main}
}

\clearpage
\appendix
\maketitlesupplementary

\section{Implementation details}
For implementation, we referenced the official source code of Delta Denoising Score\footnote{\url{https://github.com/google/prompt-to-prompt/blob/main/DDS_zeroshot.ipynb}} by using Stable Diffusion v1.4. We extracted intermediate features from self-attention layers and apply PatchNCE loss similar to CUT. Inspired by the analysis of CUT, we applied PatchNCE to all the up-sampling self-attention layers but excluded features from the U-Net bottleneck layer, as it is related to the overall semantics of the images. For hyperparameters, we utilizes patch sizes of $1\times 1$ or $2\times 2$ with 256 patches and set the weight of $\ell_{con}$ to 3.0. Other settings regarding to DDS, including the number of optimization steps, optimizer, and learning rate, adhere to the default configurations provided in the official code. The code is available to public on \url{https://github.com/HyelinNAM/CDS}. All image manipulations were conducted using an NVIDIA RTX 6000, and the processing time for editing each image was approximately 2 minutes and 50 seconds. 

\begin{table}[t]
\centering
\resizebox{0.8\columnwidth}{!}{
\begin{tabular}{c c c c}
\toprule
\multicolumn{1}{c}{\multirow{2}{*}{\textbf{Method}}} & \multicolumn{3}{c}{\textbf{Metric}} \\
\cmidrule{2-4} 
\multicolumn{1}{c}{} & CLIP Acc ($\uparrow$) & Dist ($\downarrow$) & LPIPS ($\downarrow$)\\
\midrule
{SDEdit + word swap} & {99.2$\%$} & {0.066} & {0.126} \\
{DiffuseIT} & {99.2$\%$} & {0.066} & {0.255} \\
{DDPM inv + P2P} & {85.7$\%$} & {0.073} & {0.147} \\
{DDPM inv + PnP} & {86.1$\%$} & {0.078} & {0.166}\\
{DDS} & {\textbf{100}$\%$} & {0.031} & {0.091}\\
{\textbf{Ours}} & {\textbf{{100}}}$\%$ & {\textbf{0.027}} & \textbf{{0.088}}\\
\bottomrule
\end{tabular}
} 
\caption{Quantitative evaluation for the cat $\rightarrow$ pig task. 'Dist' denots DINO-ViT structure distance.}  
\label{table: quantitative evaluation2}
\vspace{-0.3cm}
\end{table}

\section{Quantitative Results}
\begin{figure}[t]
    \centering
    \includegraphics[width=0.9\linewidth]{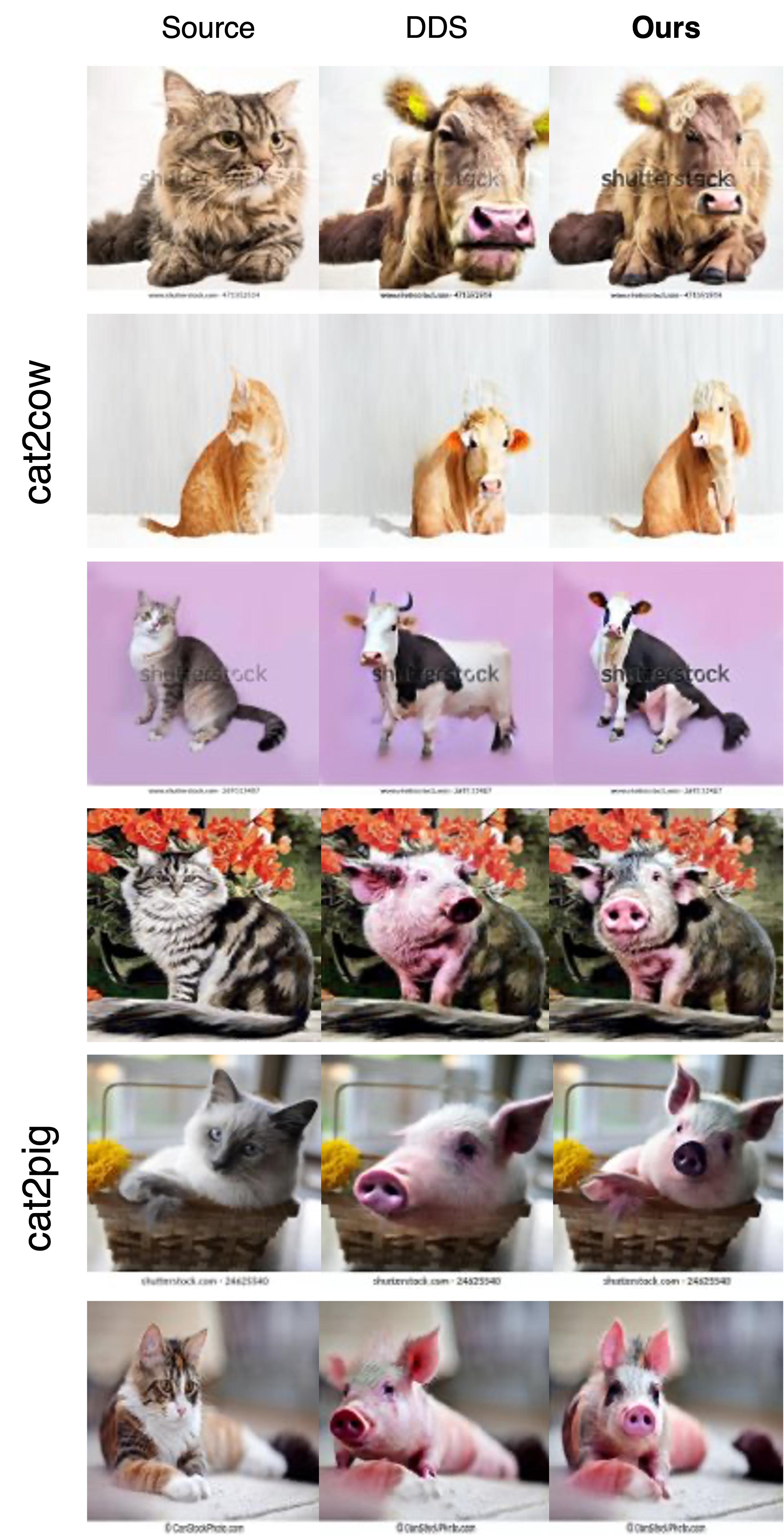}
    \caption{Sample results of the cat2dog task from DDS and CDS.}
    \label{fig:quantitative_comparison2}
    \vspace{-0.1cm}
\end{figure}

In addition to the quantitative results presented in the main paper, \cref{table: quantitative evaluation2} also demonstrates the effectiveness of our proposed method in achieving optimal editing outcomes while maintaining the structural elements of the source image. In contrast, other baseline models achieve low structure distance. 
This implies that despite achieving high CLIP accuracy, the existing methods edited the image without considering the source structure. The visual comparison for both tasks(see~\cref{fig:quantitative_comparison2}) also confirms the aforementioned quantitative evaluation.

\section{Additional Ablation studies}
\subsection{Patch size}
First, we evaluate the impact of patch size by varying its size. As shown in \cref{fig:patch_ablation}a, we observe that the patch size has an effect on the extent of content preservation. As we are regulating the latent, which is more compact than image pixels, utilizing small patch size shows better impact on preserving structural elements and background details. Therefore, we decide to use a small patch, specifically $1\times 1$ and $2\times 2$, to align with our objectives.

\subsection{The number of patches}
\begin{figure}[t]
    \centering
    \includegraphics[width=1.0\linewidth]{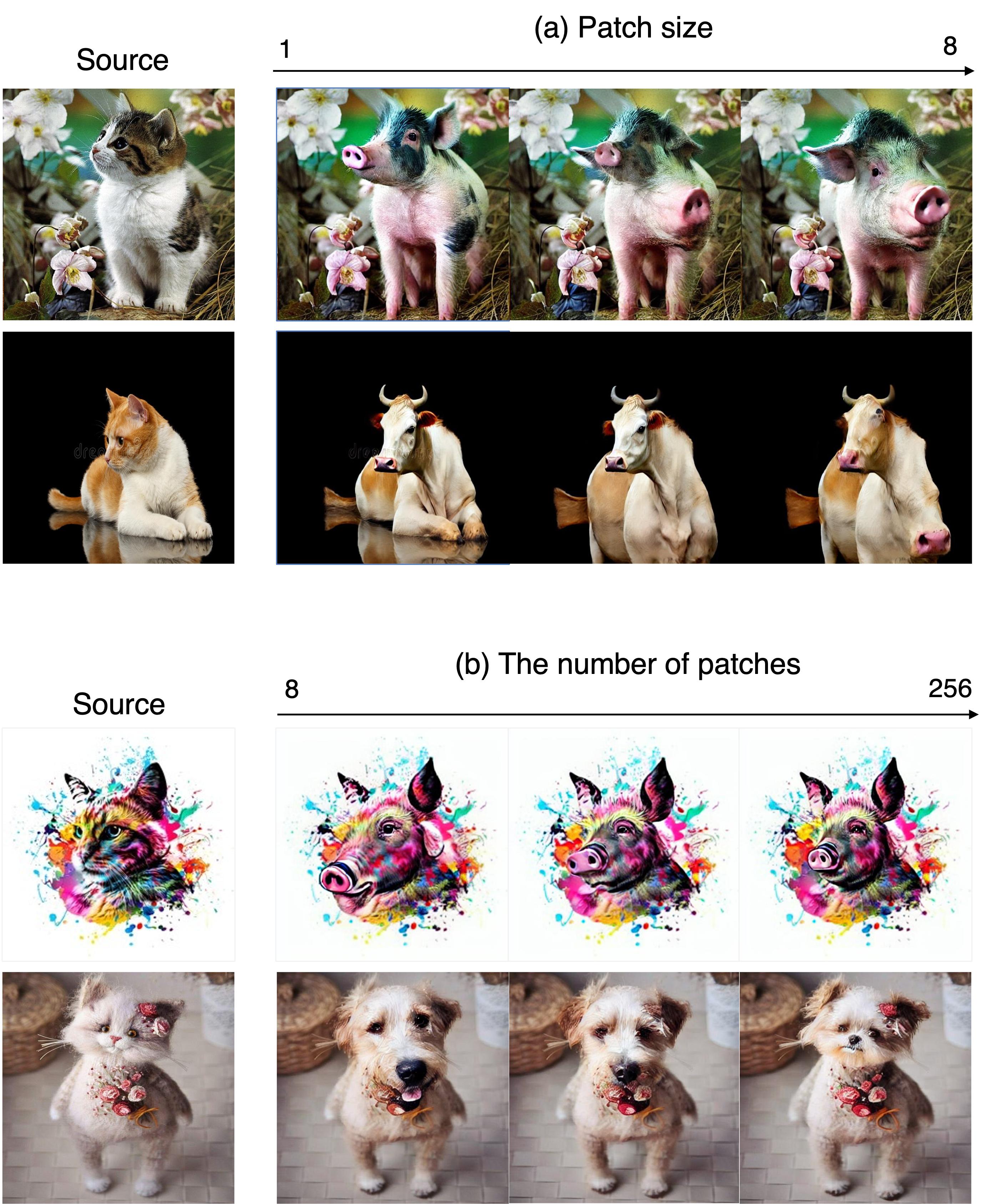}
    \vspace{-0.7cm}
    \caption{Ablation study on (a) patch size and (b) the number of patches. The given prompt is ``cat → pig," ``cat → cow,", ``cat → pig" and "cat → dog"  from top to bottom.}
    \label{fig:patch_ablation}
\end{figure}

We also ablate the impact of the number of patches and found that it also determines the extent of the regulation. As the number of patches increases, we observe a better preservation of structural aspects of the original image, such as facial structure and head angles (see \cref{fig:patch_ablation}b). Therefore, we chose 256 number of patches.

\section{Additional results}
\begin{figure}[t]
    \centering
    \includegraphics[width=0.9\linewidth]
    {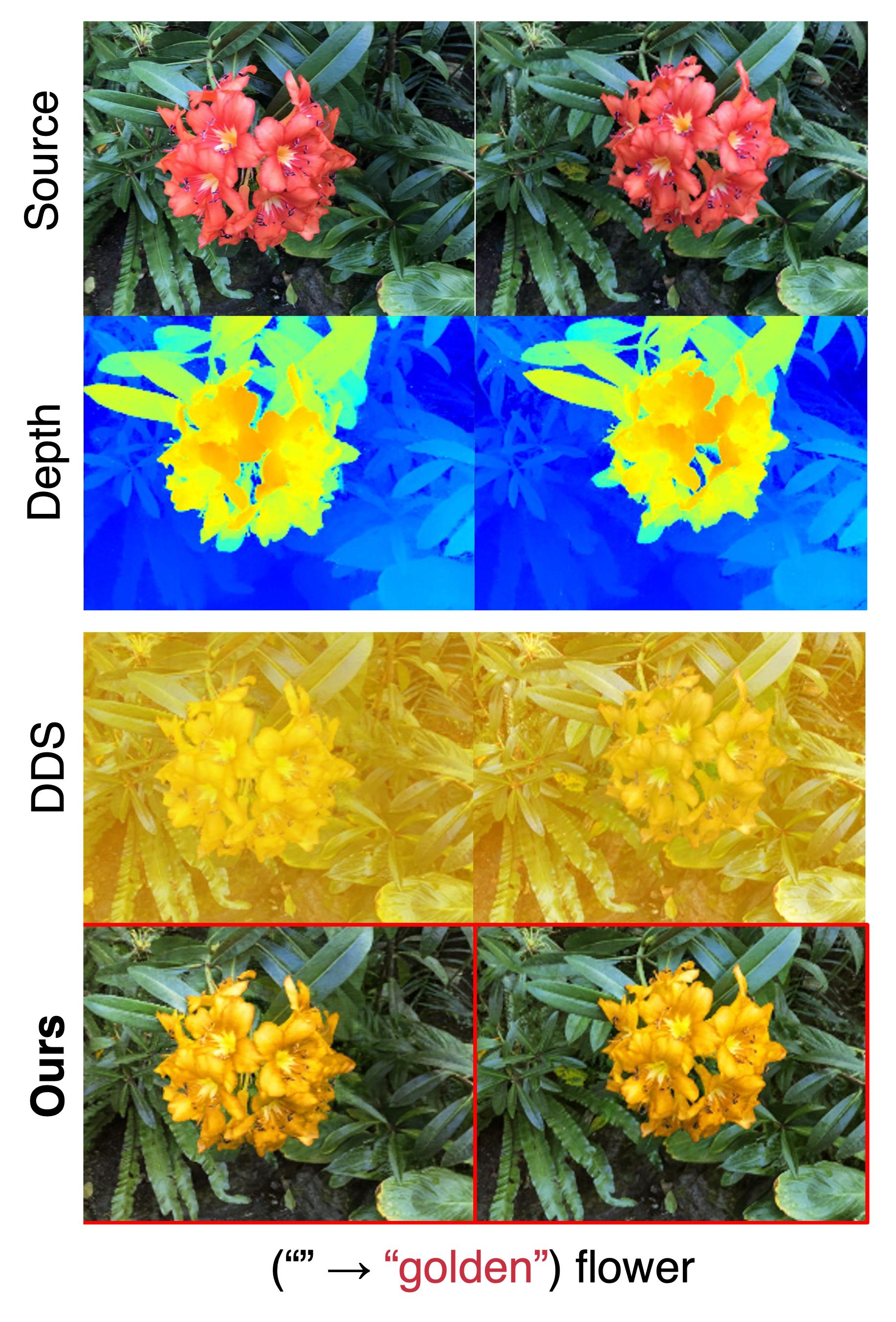}
    \vspace{-0.5cm}
    \caption{Additional results on NeRF editing.}
    \label{fig:additional_nerf}
\end{figure}

\subsection{Qualitative results}
\begin{figure*}[t]
    \centering
    \includegraphics[width=1.0\linewidth]
    {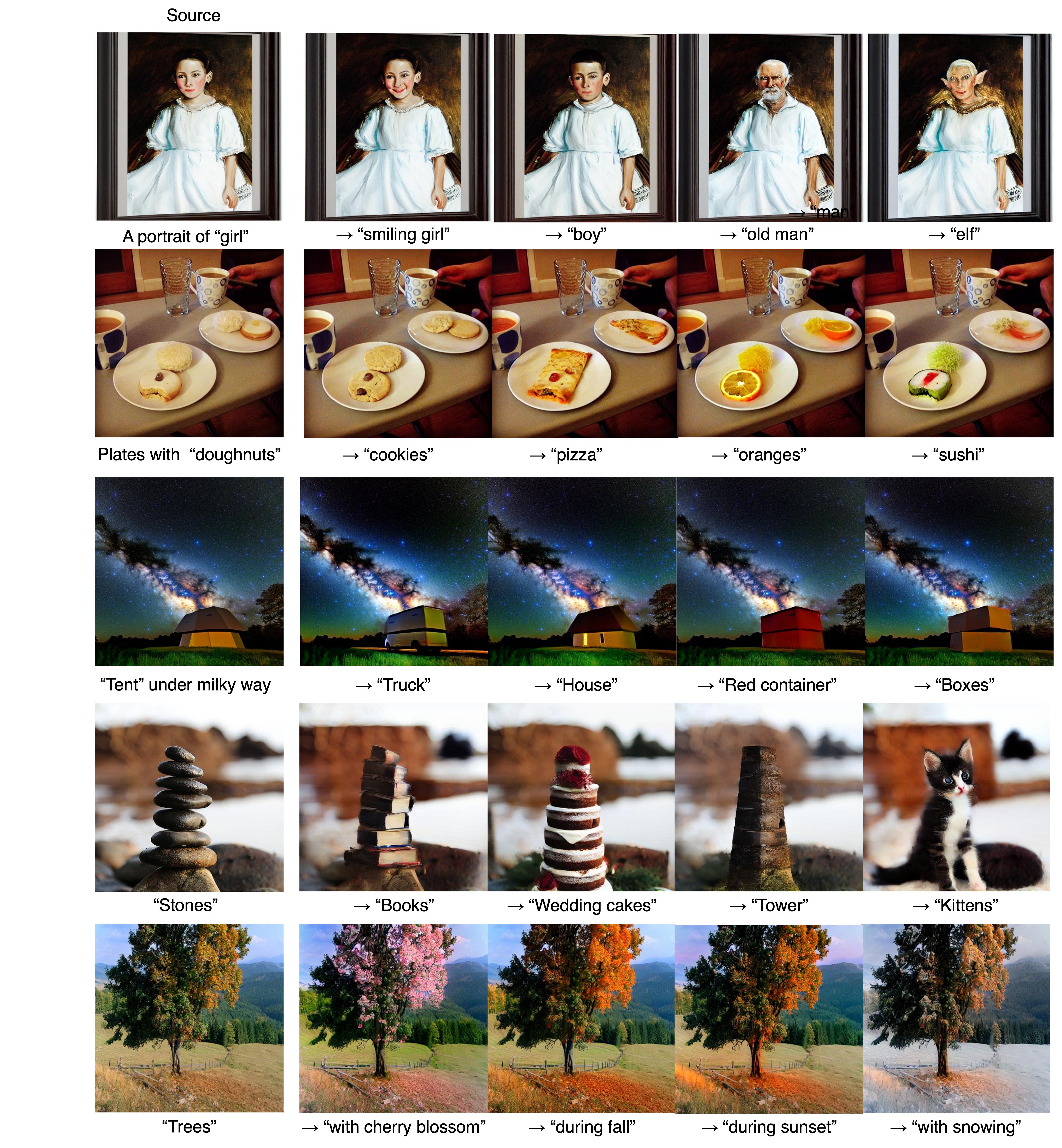}
    \vspace{-0.5cm}
    \caption{Additional qualitative results with various images and prompts.}
    \label{fig:additional_results1}
\end{figure*}

\begin{figure*}[t]
    \centering
    \includegraphics[width=1.0\linewidth]
    {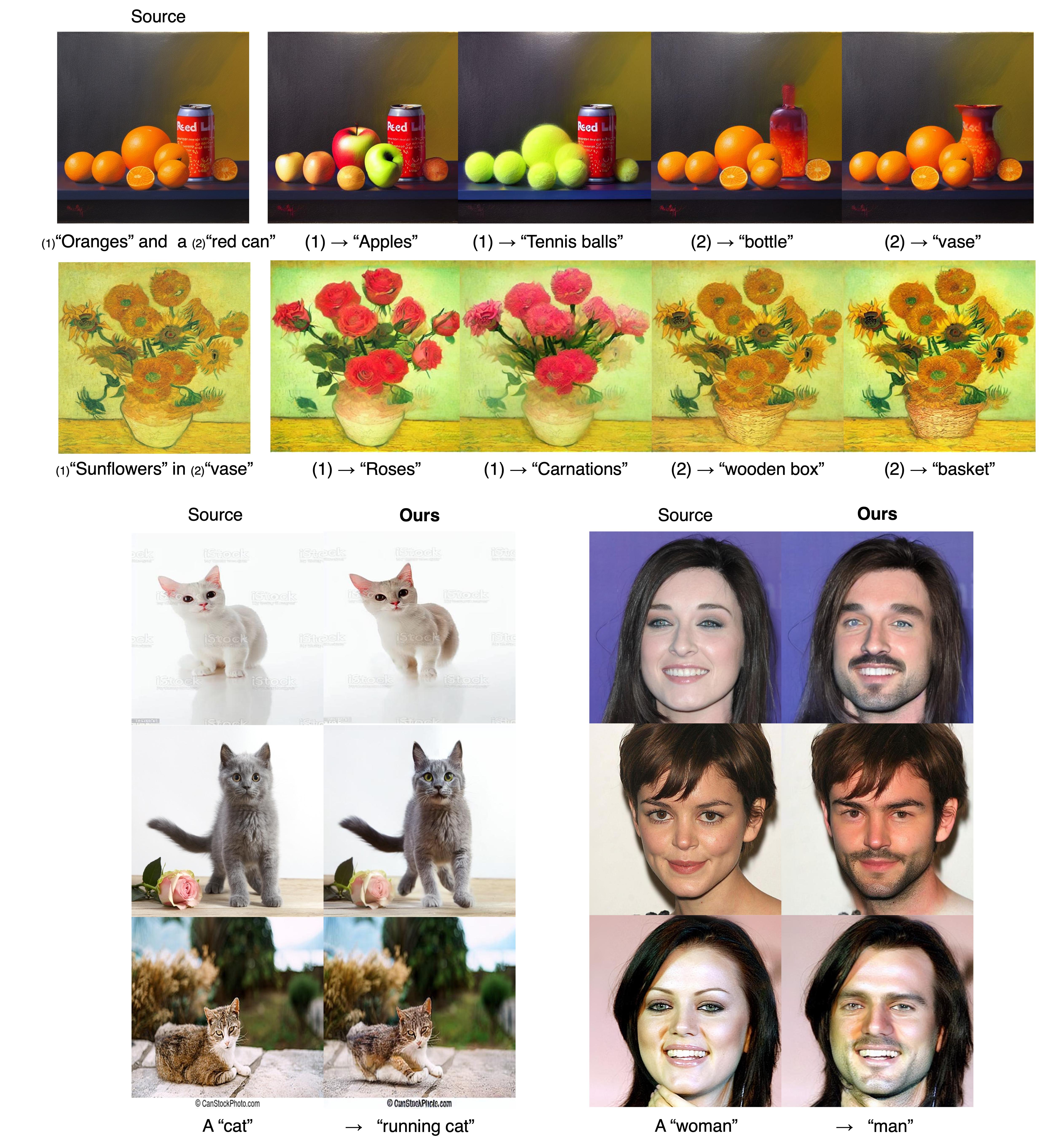}
    \vspace{-0.5cm}
    \caption{Additional qualitative results with various images and prompts.}
    \label{fig:additional_results2}
\end{figure*}

In \cref{fig:additional_results1,fig:additional_results2}, we show our edited outputs with various images and prompts. The results clearly demonstrate that our method can be applied not only to changing objects but also to diverse cases, such as adding a smile or altering gender. The proposed framework is capable of performing the edits while still retaining the other details, such as background details.

\subsection{NeRF editing}
We further provide additional results on NeRF editing. To fine-tune the NeRF model, we utilized the recent pre-trained model, TensoRF~\cite{tensorf}. For efficiency, we downsampled the images to a resolution of 504x378 in pre-training stage. For fine-tuning the pre-trained model, we further downsampled the resolution to 252x189 due to resource constraints.

For fine-tuning, we rendered source and target images from the pre-trained source NeRF model $\phi$ and fine-tuned model $\theta$, respectively. With the same view direction $d$, we obtain two rendered view $\hat{x}, x$, representing rendered 2D images from source and target model, respectively. By embedding these two images into the encoder of the Stable Diffusion model, we obtain source and target latent $\hat{\zb},\zb$.
With the prepared latents, we can calculate DDS gradient as follows:
\begin{equation} \label{eq:dds}
\mathcal{L}_{\mathrm{DDS}}(\theta,y_{trg})=\nabla_\theta \mathcal{L}_{\mathrm{SDS}}(\zb, y_{trg})-\nabla_\theta \mathcal{L}_{\mathrm{SDS}}(\hat{\zb}, y_{src}).
\end{equation}
We also utilize our proposed contrastive loss along with the aforementioned DDS gradient to update the NeRF parameter $\theta$. During training, we used the Adam optimizer with a learning rate of 0.01, and conducted 400 iterations for fine-tuning. The overall process takes about 8 minutes per sample.  

In Fig. \ref{fig:additional_nerf}, we show the comparison results between the baseline DDS and CDS. We observed that basic DDS model struggles to accurately capture the shape of the original 3D object, often resulting in changes to the entire color tones. In contrast, our method enables localized object editing without compomising the shape of the original object.

\section{Limitations and Negative social impact}
Given that our framework manipulates images based on user intentions, there is a potential for misuse, including the creation of deepfakes or other forms of disinformation. Additionally, our method has dependency on generative priors of a large text-to-image diffusion models, which may contain undesired biases. Therefore, ensuring ethical implementation and appropriate regulation are imperative for these methods.

\clearpage

\end{document}

%% file: preamble.tex
\usepackage[dvipsnames]{xcolor}
